\documentclass{article}

\usepackage{PRIMEarxiv}

\usepackage[utf8]{inputenc} 
\usepackage[T1]{fontenc}    
\usepackage[colorlinks=true, linkcolor=black, citecolor=black, urlcolor=black]{hyperref}
\usepackage{url}            
\usepackage{booktabs}       
\usepackage{amsfonts}       
\usepackage{nicefrac}       
\usepackage{microtype}      
\usepackage{lipsum}
\usepackage{fancyhdr}       
\usepackage{graphicx}       
\graphicspath{{media/}}     

\title{

KIRETT - A wearable device to support rescue operations using artificial intelligence to improve first aid

\thanks{\textit{2022 IEEE International Smart Cities Conference (ISC2) | 978-1-6654-8561-6/22/\$31.00 ©2022 IEEE | DOI: 10.1109/ISC255366.2022.9922361.}
}}

\author{
  Johannes Zenkert, Christian Weber, Mubaris Nadeem, Lisa Bender, Madjid Fathi \\
  Institute for Knowledge-Based Systems and Knowledge Management \\
  University of Siegen\\
  Hoelderlinstrasse 3, 57068 Siegen, Germany\\
  \texttt{\{Johannes Zenkert\}johannes.zenkert@uni-siegen.de} \\
   \And
  Abu Shad Ahammed, Aniebiet Micheal Ezekiel, Roman Obermaisser, Maximilian Bradford \\
  Chair for Embedded Systems\\
  University of Siegen\\
  Hoelderlinstrasse 3, 57068 Siegen, Germany\\
  57068 Siegen, Germany
}

\begin{document}
\maketitle

\begin{abstract}
This short paper presents first steps in the scientific part of the KIRETT project, which aims to improve first aid during rescue operations using a wearable device. The wearable is used for computer-aided situation recognition by means of artificial intelligence. It provides contextual recommendations for actions and operations to rescue personnel and is intended to minimize damage to patients due to incorrect treatment, as well as increase the probability of survival. The paper describes a first overview of research approaches within the project.
\end{abstract}


\section{Introduction}
Rescue personnel worldwide are under extreme pressure and have to make decisions in the shortest possible time which can make a difference between life or death. For them, it is necessary to recognize relevant health situations of emergency patients on site and to take appropriate first aid measures. It happens often that an emergency patient is facing drastic changes in health vitals for an unrecognized reason, which results in certain modifications of emergency treatments. That is why it is necessary to find a fast and reliable solution for assessing the patient's condition in real time and provide necessary treatment in rescue scenarios. The 'KIRETT' project is developed to support the rescue personnel by means of a wearable that will be used for situation recognition, context-based treatment recommendations and minimizing potential damage due to incorrect treatments, to increase the probability of survival and lower the risk of long time impairments. To improve the quality of care and to increase the efficiency of rescue operations, an artificial neural network (ANN) algorithm will be developed to recognize the patient's health situation. Afterwards, using a knowledge graph, the proper treatment path will be advised to the rescue personnel through the wearable.\\
ANN is often used by current researchers to identify health complications like diabetes, cancer, asthma or cardiac arrest. AI based disease detection algorithms proved to be an effective tool for correctly diagnosing these health complications and creating numerous opportunities for reducing human errors, improving clinical outcomes, tracking data over time, etc. \cite{r1}. In the given research study, the developed ANN model will provide a vector output indicating the probability of complications that can exist when a rescue attempt is initiated. The model will be trained with records of the past five years from the rescue station from Siegen-Wittgenstein in Germany. Before the training, the raw dataset from the rescue station will be sorted and filtered using open-source data management system and Python programming. All the data modifications and analysis tools development including knowledge graph is done in local computer.
The rescue station will also provide the literature of treatment instructions that consists of standard operating procedures in rescue services, as well as the guidelines for treatment and expert knowledge. Text mining will be used as an analysis method to extract information from the literature and other structured and unstructured knowledge sources. The extracted information will be stored in the knowledge base as dimensions, which can be used among other things, for the construction of knowledge graphs to display recognized connections or linked content or as a basis for the inference of new knowledge.
\section{Situation recognition through neural network architecture}
A neural network will be used to recognize likely emergency health situations e.g., cardiac arrest, respiratory disease, pulmonary embolism of a patient. The ANN algorithm will be trained with more than 300,000 past rescue records consisting of health vital information, the primary diagnosis, first aid plans and rescuers' impressions. In the further course of patient's treatment, the situation may change afterward as the rescue forces assessing this primarily on the basis of the visible condition of the emergency patient, the data from the medical equipment (e.g. ventilator, ECG) and the mission description from the control center. For example, complications such as a reduction in respiratory drive during ventilation or cardiac arrhythmia during a heart attack can occur. Such situations must be responded to immediately, for example by resuscitation or appropriate medication. The developed neural network will help to make a quick decision and save the patients by continuously interpreting their situation based on health vitals and access to more data sources as soon as the equipment is attached in first aid situation. 
\subsection{Identification of the complications and their attributes}
The first step in developing the neural network is to create a training data set which consists of multiple output labels. The labels represent the complicated disease groups that can occur in an emergency situation. In collaboration with the rescue station, Siegen and professional medical partners, the possible complication groups to be labelled as ANN output are identified. From the rescue station database, it is found that the most common complications occur in emergency situations are pulmonary diseases, CNS diseases, cardiovascular diseases, respiratory diseases, abdominal diseases, psychiatric diseases, metabolic diseases, gynecologic-obstetrical emergencies, infections, and some other special diseases. For each complication group, such as cardiovascular and neurological disease, a thorough research was conducted to find out the most relevant attributes which are most common as disease symptoms and can be used to identify the complications. The attributes are mostly health vitals like blood oxygen saturation (SpO2), systolic and diastolic blood pressure, blood sugar level etc. Furthermore, text attributes containing important health information or diagnostic information from rescue personnel are categorized based on the severity, disease relevance and expert opinion.\\
The complication groups are composed of multiple diseases. The current approach is to first identify the patient's situation with a particular complication group, but not a particular disease that falls under that group. This is due to the number of total diseases inside all those groups being large in quantity and if used as output labels for the ANN may cause overfitting and high output latency.
\subsection{Data sorting and filtering}
The database of past rescue attempts consists of  thousands of unique cases identified with numerous complication types. Each of those cases was primarily recorded by the first responder at the rescue station with the relevant important information: Parameters of the rescue, geographical information of the operation, patient's medical history, health diagnosis, vital signs, medications administered and treatment path performed. The database handed over for research was in raw form. Due to the manual documentation by the experts, not every case was provided with sufficient or complete information. Furthermore, much of the information recorded in the database was irrelevant for later analysis. So, a filtering algorithm is created to limit information, remove outliers and extrapolate some data fields when further knowledge sources are available.\\
A data dictionary software is created that makes a comprehensive analysis of the nature of the selected attributes and supports an on-demand data composition and export. It furthermore stores descriptions of all acronyms and data types to support the collaboration between data engineers. The software is written using Python version 3.9 and the user interface is developed using the Python library "tkinter". This software is used for data preparation and training of the neural network for situation recognition and is not used directly on the wearable. In the next phase, case IDs relevant to each complication group are created. The goal is to divide the cases based on the first impressions of the ambulance staff during a rescue mission. Several methods are used to make this initial subdivision and assign cases to a specific complication group. Accordingly, an analysis is done to sort out the diagnostic information associated with each of these cases.

\subsection{Smart data analysis}
A software is developed, using Python, to merge all the records provided by the rescue station in a single file. Previously, the rescue history was recorded in more than 70 files with duplicate information. A merging of the records is performed, considering the case IDs as master keys. To explore the data constructively, an open source object-relational database management system (ORDBMS) like PostgreSQL is used. This database management system can serve multiple clients \cite{r3} and exhibit higher quality of performance when the environment requires a precise and structured data model \cite{r4}. The data received from the rescue station, originally stored in .csv format, is migrated to a PostgreSQL database using the client pgAdmin 4- version 6.7. The next task is to encode the data categorically, which is a requirement for creating the training dataset. Non-hierarchical data that can't be encoded based on category, like whether a patient had bronchitis, pneumonia or COPD, are one hot encoded. One hot encoding is a method that converts each word in a text by a numerical positional vector whose elements are all zeros, except for the position of the word in the vocabulary list \cite{r5}. After the categorization and encoding, the text columns are parsed using the TF-IDF (term frequency – inverse document frequency) vectorizer, an algorithm used to extract attributes or features from an uncategorized document to describe it efficiently. The logic behind TF-IDF is that a word becomes more relevant or essential if it occurs multiple times in a single or few documents. But when a word appears multiple times in all or the majority of the documents, then this term is considered trivial \cite{r6}. This is intended to find words that are used often but not constantly, which should statistically contain the most information.

\section{Analysis of standard operating procedures and treatment paths}
The second part of the project is used to transfer, elaborate and define rescue operation actions and instructions into the form of a knowledge graph, as a base to be able to extract and complete possible paths of action. In the first step, the standard operating procedures of basic medical measures and specialized treatment paths provided by the Siegen Rescue Station \cite{BPR} were analyzed using a text mining framework. Text content was extracted from a PDF document provided and transferred to a knowledge base. Various pre-processing steps of text analysis and the MKR (Multidimensional Knowledge Representation) framework were initially used to build a knowledge base \cite{r2}. Based on the initial knowledge base created by MKR, the representation format has been used for the modeling of the knowledge graph. Here, for example, recognized entities of the standard operating procedures contained in the text of the description were already validated with Linked Open Data.

\subsection{Definition of initial use cases}
Based on the standard operating procedures and treatment paths, use-cases were defined in several sessions with medical experts. It was initially agreed to focus on the acute coronary syndrome (ACS). The associated treatment paths were discussed in detail together with the advance trauma life support \cite{carmont_advanced_2005} measures preceding them, modeled as graphs, and all possible processes were discussed from a technical and professional perspective.

\subsection{Neo4j knowledge graph}
The treatment paths and actions of the rescue operations are modelled in a Neo4j graph database. Based on the provided medical literature for rescue operations and guidelines, the modeling, in style of an event-driven process chain (EPC), was reviewed, made available as MKR representation, and then transferred into the Neo4j graph database. The graph database is stored in the wearable memory.

For the system architecture, a component-based implementation is used, through which messages can be exchanged between components and, for example, the graph database can be queried for the next action instruction. The graph component uses interfaces to other components and sends messages for this purpose, such as the query of a data value from the internal database by the middleware, in order to then display it on the touchscreen via the visualization component after rule execution of the rules contained in the graph.
An accomplishment of such a system is possible with the use of an interface, i.e. display with touch function, to visualize components and to analyze given inputs for further steps through the graph database.

\subsection{Python implementation of the interface for the output of possible action instructions}
The language Python is used on the wearable for the implementation and to send queries to the graph database for possible actions. Here, a flexible algorithmic approach was chosen that supports the later addition of further nodes and edges in the graph. With the output of the situation detection component, a position determination in the knowledge graph is aimed at in order to determine the initial start position of the treatment graph. For this purpose, the output of the neural network is processed rule-based and the recognized disease group with the highest probability is used for a potential start node from the corresponding treatment paths. Results are displayed on the touchscreen in order to manually interfere or adjust by the rescue personnel.
For development, a locally provided Jupyter notebook, connected to the local graph database was used. The developments are transferred to the software environment on the wearable.

\subsection{Communication between neural network and knowledge graph}
First approaches have been developed for the communication between the neural network, which is to describe the current situation in the emergency operation, and the treatment paths that are selected based on the situation. It is possible to have several starting points for the treatment within the knowledge graph of the graph database. The neuronal network provides a distribution of probabilities of all possible situations as an output (e.g., disease group patterns recognized by data values and information from the control center, including classification of the operation). The collected probabilities of situations at a given time and detected by the neural network, are used by the main application to query the graph database for the most likely starting point. Situation detection runs continuously in the background during the deployment, incorporating newly acquired data (e.g. temperature measurement, ECG, etc.) and changing the probability for starting points. Via the main application, this will be communicated to the graph database interface, showing the most likely treatment path as well as further probable paths to change to as a graphical representation on the touch display. Thus, possible information unnoticed during the rescue operation can be communicated to the rescue personnel and a suggestion for another possible treatment path can be communicated while the operation is still in progress.

\section{Portable device with energy efficiency, real-time capability and reliability}
\subsection{Wearable implementation}
\begin{figure}[h]
\centerline{\includegraphics[width=8.9cm]{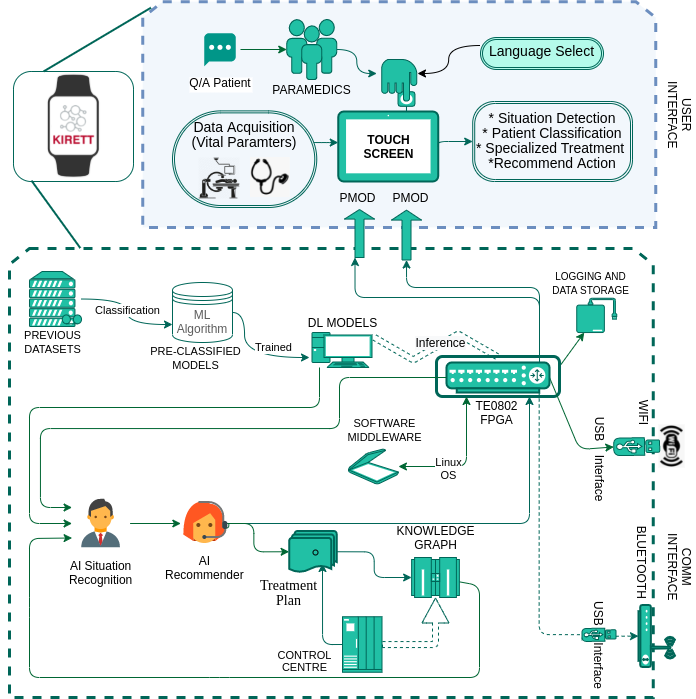}}
\caption{Hardware Implementation of the Wearable}
\label{fig1}
\end{figure}

A system architecture was defined with hardware and software integration to create a unified device. An implementation of the wearable device is elaborated as shown in Figure \ref{fig1}. The implementation includes inputs to the wearable: a) language selection, b) acquisition of time series of vital signs (from medical devices via WIFI or Bluetooth), c) information from the control center and d) inputs from paramedics (question/answer dialogues via touchscreen). The outputs on the wearable screen are: a) recognition of the situation, b) classification of patient severity, c) a specialized treatment plan, and d) recommended actions.

\subsection{Apache TVM/VTA Deep Learning Compiler Stack and Hardware Accelerator}
The shift towards hardware specialization and the growing interest in open-source hardware platforms for running machine learning (ML) and deep learning (DL) applications prompted the choice of Apache TVM as the toolchain to deploy the DL model of medical complications into an edge FPGA. The project prioritizes security and network stability, and the local device used maximizes the handling of data privacy \cite{d1}, a key criterion during rescue missions. The Apache TVM is a modern operating system platform for ML models with an end-to-end TVM stack integration flexible for our applications. It allows direct optimization of DL models, and its front-end supports most deep learning frameworks: TensorFlow, PyTorch, Keras, MXNet, et. cetera. The trained model is accelerated on the FPGA using the VTA (versatile tensor accelerator), an open-source Deep Learning accelerator that complements the end-to-end TVM-based compiler stack. The VTA is generic and enables open-source hardware with an optimized workflow for deploying models on FPGAs. To support the FPGA hardware backend, TVM/TVA has preset drivers, a JIT runtime responsible for generating an instruction set for the hardware abstraction layer \cite{d2}. In addition, the VTA flexible RPC interface facilitates hardware deployment of the trained models and allows the programming of the chosen FPGA with the benefits of Python.

\subsection{Apache TVM/VTA Extension: Reliability, Predictability, and Fault Tolerance}
In domain-specific embedded systems such as those in the medical domain, AI applications take into account safety critical mechanisms to avoid catastrophic outcomes in events of failure \cite{d3}. However, TVM/VTA is not flawless in its AI implementation, requiring evaluation of its reliability in safety critical systems \cite{d4} as it is prone to significant uncertainty. Therefore, for the suitability of this embedded AI platform for the rescue mission, Apache TVM/VTA was extended to provide temporal predictability. Furthermore, this state-of-the-art approach combines fundamental reliability-related concepts in terms of safety, redundancy, resource efficiency, and fault tolerance \cite{d5} \cite{d6}.

\section{Conclusion}
This paper summarizes the initial scientific results of the KIRETT project and describes the progress of the key activities carried out for introducing a situation recognition and knowledge graph based recommendations for action. In the future, the project will be focused on the realization of the wearable and the integration of individual components.

\section*{Funding}
The ongoing research was financially supported by the Federal Ministry of Education and Research, Germany. 
The research has been supported by KIRETT project coordinator CRS Medical GmbH (Aßlar, Germany), and partner organization mbeder GmbH (Siegen, Germany). The authors would like to thank the associative partners of the project: Kreis Siegen-Wittgenstein, City of Siegen, the German Red Cross Siegen (DRK) and the Jung-Stilling-Hospital in Siegen.

\bibliographystyle{unsrt}
\bibliography{templateArXiv}

\begin{thebibliography}{10}

\bibitem{r1}
Yogesh Kumar, Apeksha Koul, Ruchi Singla, and Muhammad~Fazal Ijaz.
\newblock Artificial intelligence in disease diagnosis: a systematic literature
  review, synthesizing framework and future research agenda.
\newblock {\em Journal of Ambient Intelligence and Humanized Computing}, pages
  1--28, 2022.

\bibitem{r3}
I.~S. Vershinin and A.~R. Mustafina.
\newblock {Performance Analysis of PostgreSQL, MySQL, Microsoft SQL Server
  Systems Based on TPC-H Tests}.
\newblock In {\em 2021 International Russian Automation Conference
  (RusAutoCon)}, pages 683--687. IEEE, 2021.

\bibitem{r4}
Min-Gyue Jung, Seon-A Youn, Jayon Bae, and Yong-Lak Choi.
\newblock A study on data input and output performance comparison of mongodb
  and postgresql in the big data environment.
\newblock In {\em 2015 8th international conference on database theory and
  application (DTA)}, pages 14--17. IEEE, 2015.

\bibitem{r5}
Joseph~D Prusa and Taghi~M Khoshgoftaar.
\newblock Improving deep neural network design with new text data
  representations.
\newblock {\em Journal of Big Data}, 4(1):1--16, 2017.

\bibitem{r6}
Gleen~A. Dalaorao, Ariel~M. Sison, and Ruji~P. Medina.
\newblock Integrating collocation as tf-idf enhancement to improve
  classification accuracy.
\newblock In {\em 2019 IEEE 13th International Conference on Telecommunication
  Systems, Services, and Applications (TSSA)}, pages 282--285, 2019.

\bibitem{BPR}
{Kreis Siegen-Wittgenstein - Arbeitsgruppe BPR/SAA}.
\newblock {Behandlungspfade und Standardarbeitsanweisungen für den
  Rettungsdienst im Kreis Siegen-Wittgenstein}, {October} {2020}.

\bibitem{r2}
Johannes Zenkert, André Klahold, and Madjid Fathi.
\newblock Knowledge discovery in multidimensional knowledge representation
  framework.
\newblock {\em Iran Journal of Computer Science}, 1(4):199--216, 4 2018.

\bibitem{carmont_advanced_2005}
M.~R. Carmont.
\newblock The advanced trauma life support course: a history of its development
  and review of related literature.
\newblock 81(952):87--91.
\newblock Publisher: The Fellowship of Postgraduate Medicine Section: Review.

\bibitem{d1}
Taehee Jeong, Ehsam Ghasemi, Jorn Tuyls, Elliott Delaye, and Ashish Sirasao.
\newblock Neural network pruning and hardware acceleration.
\newblock In {\em 2020 IEEE/ACM 13th International Conference on Utility and
  Cloud Computing (UCC)}, pages 440--445. IEEE, 2020.

\bibitem{d2}
Ravikumar~V Chakaravarthy and Hua Jiang.
\newblock Special session: Xta: Open source extensible, scalable and adaptable
  tensor architecture for ai acceleration.
\newblock In {\em 2020 IEEE 38th International Conference on Computer Design
  (ICCD)}, pages 53--56, 2020.

\bibitem{d3}
Ryan Luna and Sheikh~Ariful Islam.
\newblock Security and reliability of safety-critical rtos.
\newblock {\em SN Computer Science}, 2, 09 2021.

\bibitem{d4}
Lucas Matana~Luza, Annachiara Ruospo, Alberto Bosio, Ernesto Sanchez, and Luigi
  Dilillo.
\newblock A model-based framework to assess the reliability of safety-critical
  applications.
\newblock pages 41--44, 04 2021.

\bibitem{d5}
Roman Obermaisser, Hamidreza Ahmadian, Adele Maleki, Yosab Bebawy, Alina Lenz,
  and Babak Sorkhpour.
\newblock Adaptive time-triggered multi-core architecture.
\newblock {\em Designs}, 3:7, 01 2019.

\bibitem{d6}
Alessandro Biondi, Federico Nesti, Giorgiomaria Cicero, Daniel Casini, and
  Giorgio Buttazzo.
\newblock A safe, secure, and predictable software architecture for deep
  learning in safety-critical systems.
\newblock {\em IEEE Embedded Systems Letters}, 12(3):78--82, 2020.

\end{thebibliography}

\end{document}